\documentclass[runningheads]{llncs}
\pdfoutput=1 

\usepackage[numbers,sectionbib]{natbib}
\usepackage{amssymb}
\usepackage{bbm}
\usepackage{amsmath}
\usepackage{amsfonts}
\usepackage{mathtools}

\usepackage{dirtytalk}

\DeclareMathOperator{\Prec}{Prec}
\DeclareMathOperator{\cov}{cov}

\newcommand{\abs}[1]{\left\lvert#1\right\rvert}  
\DeclareMathOperator*{\argmax}{arg\max}

\newcommand{\defeq}{\vcentcolon =}

\newcommand{\card}[1]{\left\lvert#1\right\rvert}
\newcommand{\condexpec}[2]{\mathbb{E}\left[#1\mid #2\right]}
\newcommand{\corpus}{\mathcal{C}}
\newcommand{\dictionary}{\mathcal{D}}
\newcommand{\localdic}{\mathcal{D}}
\newcommand{\mult}{m}
\newcommand{\indicator}{\mathbbm{1}}
\newcommand{\precision}[1]{\Prec\left(#1\right)}
\newcommand{\Reals}{\mathbb{R}}
\newcommand{\word}{w}
\newcommand{\jaccard}[2]{J\left(#1,#2\right)}
\newcommand{\logiwords}{\Lambda}
\newcommand{\doc}{z}

\usepackage[hyphens]{url}

\usepackage{diagbox}

\usepackage{graphicx}
\usepackage{xcolor}

\begin{document}

\title{Comparing Feature Importance and Rule Extraction for Interpretability on Text Data}

\titlerunning{Comparing Interpretability on Text Data}
%
\author{Gianluigi Lopardo
\and
Damien Garreau
}
\authorrunning{G. Lopardo and D. Garreau}

\institute{Université Côte d'Azur, Inria, CNRS, LJAD, France 
}

\maketitle 

\begin{abstract}
Complex machine learning algorithms are used more and more often in critical tasks involving text data, leading to the development of interpretability methods. 
Among local methods, two families have emerged: those computing importance scores for each feature and those extracting simple logical rules. 
In this paper we show that using different methods can lead to unexpectedly different explanations, even when applied to simple models for which we would expect qualitative coincidence. 
To quantify this effect, we propose a new approach to compare explanations produced by different methods. 

\end{abstract}


\section{Introduction}
%
In recent years, increased complexity seems to have been the key to obtain state-of-the-art performance in natural language processing. 
Language models such as BERT \citep{devlin_et_al_2018} or GPT-3 \citep{Brown_et_al_2020} typically rely on billions of parameters and complex architecture choices to make accurate predictions. 
The availability of huge datasets and the computational capacity available today make this growth in complexity of algorithms possible. 
On the other hand, the opacity of these models hinders their usage in sensitive domains, such as healthcare or legal.

Indeed, there is a lack of adequate explanations to support individual predictions, preventing the social acceptance of these decisions. 
In order to provide interpretability, numerous methods have been proposed in the last five years~\citep{guidotti2018survey,linardatos2021explainable}. 
In this paper, we focus on local, \emph{post hoc} explanations, that is, methods explaining one decision in particular for a model which is already trained. 
There is a great diversity among these methods which we summarize briefly here. 
Perhaps the easiest to understand compute the gradient (or a variation thereof) of the model with respect to the input \citep{ancona_et_al_2018}. 
Other methods, such as LIME \citep{ribeiro2016should} and kernel SHAP \citep{lundberg2017unified} give attribution scores to each feature by fitting a linear model on the presence or absence of a feature.  
Rule-based methods such as Anchors~\citep{ribeiro2018anchors} determine a small set of rules satisfied by the instance and provide it as explanation. 
Their principle is to learn decision sets that jointly maximizes their interpretability and predictive accuracy~\citep{lakkaraju2016interpretable}. 
Explanations in form of rules are typically preferred by users~\citep{lim2009and}. 
Let us also mention that attention mechanisms~\citep{vaswani_et_al_2017}, more and more frequently used in deep neural networks architectures, can be leveraged to get interpretability. 

One main problem in interpretability is the lack of adequate metrics to measure the quality of explanations. 
While some studies propose a framework for comparing feature importance methods \citep{bhatt2021evaluating, nguyen2020quantitative} and others for comparing rule-based methods \citep{margot2021new}, the comparison between methods of different classes is more challenging. 
Moreover, the problem is particularly understudied on textual data. 

In this paper, we focus on perturbative and rule-based approaches, specifically LIME and Anchors.
Our goal is to pin-point differences in their results which can be easily overlooked.  
Our motivation for doing so is the following: for a user working with a specific model and instance, the results of LIME and Anchors are qualitatively similar---both will highlight a subset of the words used in the document.
It is tempting to think that these two subsets should roughly match. 
Focusing on the sentiment prediction task, we show empirically that this is not the case, even for very simple classifiers such as logistic models. 

The paper is organized as follows: we first recall briefly the methods that we are scrutinizing in Section~\ref{sec:methods}. 
We then present our main findings in Section~\ref{sec:results}, before concluding in Section~\ref{sec:conclusion}. 
The code used for the comparison is available at \url{https://github.com/gianluigilopardo/anchors_vs_lime_text}, where our experiments are reproducible.


\paragraph{Notation.}
In all the paper, we will consider a model~$f$ applied to  text documents~$\doc$ of length~$b$ ($\doc$ contains $b$ words). 
We let $d$ denote the number of unique words of $\doc$, which is potentially smaller than $b$. 
For a given corpus $\corpus$, we define $\dictionary=\{\word_1,\ldots,\word_D\}$ as the global dictionary with cardinality $D=\card{\dictionary}$, containing the distinct words of each document in $\corpus$. 
For any given document $\doc$, we can define a \emph{local dictionary} $\localdic(\doc)$, containing a subset $d$ of $\dictionary$. 
We set $\mult_j$ the multiplicity of word $j$ in $\doc$ (in particular, $\mult_j=0$ if word $\word_j$ does not appear in document $\doc$). 
Finally, for any integer $k$, we set $[k]= \{1,\ldots,k\}$.


\section{Methods}
\label{sec:methods}
In this section, we briefly recall the operation procedure of LIME (Section~\ref{sec:lime}) and Anchors (Section~\ref{sec:anchors}), introducing our notation in the process. 
Our main assumption going into that description is that the classifier~$f$ takes as input the TF-IDF \citep{luhn1957statistical} vectorization of the words. 
We denote by $\phi$ this mapping.


\subsection{LIME for text data}
\label{sec:lime}

LIME \citep{ribeiro2016should} provides explanation in the form of feature attribution for the presence or absence of a unique word in the document to explain $\xi$. 
Since our choice is set on a given vectorizer $\phi$, the procedure is as follows:
\begin{enumerate}
\item create $n$ ($=10^3$) perturbed samples from $\xi$ by removing words at random;
\item get the predictions $y_i=f(\phi(x_i))$;
\item train a weighted linear model on the presence / absence of words.
\end{enumerate}

\paragraph{Sampling.}
The sampling procedure is as follows: for each perturbed document $x_i$, draw $s_i$ a number of deletions uniformly at random in $[d]$. 
Then draw uniformly at random a subset $S_i\subseteq \localdic(x)$ of size $s_i$ and remove all corresponding words from the document. 
In particular, all occurrences of a given word selected by this procedure are removed. 

\paragraph{Surrogate model.}
Further, weights $\pi_i$ are given to each perturbed sample $x_i$.
Finally, a linear model is fitted on the $y_i$ with inputs given by the indicator functions that word $j$ belongs to $x_i$ and weights $\pi_i$. 
The user is provided with a visualization of the weights of this linear model. 


\subsection{Anchors for text data}
\label{sec:anchors}
An \textit{anchor} is defined by \citet{ribeiro2018anchors} as a logical condition that \textit{sufficiently} approximates the model locally. 
In the case of textual data, anchors are simply a subset of the words in the example~$\xi$. 
The precision of an anchor $A$ for a prediction $f(\xi)$ is defined as $\precision{A}=\condexpec{\indicator_{f(x)=f(\xi)}}{A}$, where the condition means that all words in $A$ belong to $x$.
Since $\precision{A}$ is generally not available in practice, an empirical estimate of the precision is computed from new samples $x_i$ of the~text. 

The core idea of Anchors is to pick an anchor with high precision, while preserving some notion of globality. 
More precisely, Anchors solves~(approximately)
\begin{equation}
\label{eq:condition}
A \in \argmax_{\precision{A}\geq 1-\varepsilon}{\cov(A)} 
    \,,
\end{equation}
where, by default $\varepsilon=0.05$ and the coverage $\cov(A)$ is defined as the probability that $A$ applies to samples. 
However, due to Anchors' sampling, maximizing the coverage is equivalent to minimizing the length of $A$ (see \citet{lopardo2022sea} for more details). 

\paragraph{Sampling.}
As for LIME, the idea is to look at the behaviour of the model $f$ in a local neighborhood of $\xi$, while fixing the anchor.
For a given document $\xi$ and each candidate anchor $A\subseteq\xi$, the sampling is performed in the following steps:
\begin{enumerate}
    \item a number $n$ ($=10$) of identical copies $x_1,\ldots,x_n$ of $\xi$ are generated;
    \item for each word $\xi_k$ not in $A$, any $x_{i,k}$ is selected with probability $1/2$;
    \item selected words are then \emph{removed} by replacing them with the token \say{UNK.}
\end{enumerate}
Finally, the model is queried on these samples and the empirical precision is computed. 
The user is provided with the shortest anchor satisfying the precision condition of Eq. \eqref{eq:condition} (note that it is not necessarily unique). 

One main difference between LIME and Anchors lies in the sampling. 
LIME selects words to remove in the local dictionary $\localdic_\xi$: if a word is selected, all its occurrences in $\xi$ will be removed. 
Anchors consider words in $\xi$ as independent.


\section{Main results}
\label{sec:results}
We now present our main results, comparing LIME and Anchors for text data when applied to simple classifiers. 
We run experiments on three reviews datasets: Restaurants, Yelp, and IMDB, available on Kaggle. 
We work with (binary) sentiment analysis: label $1$ denotes a positive review and $0$ a negative one.
Note that we always consider explaining positive predictions, \emph{i.e.}, we look at examples $\xi$ such that $f(\xi)=1$. 
In Section \ref{sec:qualitative}, we present a qualitative comparison of LIME and Anchors, by looking at individual explanations. 
In Section \ref{sec:quantitative} we propose the $\ell$-index: a new metric to measure the quality of explanation on text. 
Unless otherwise specified, the figures will report the average LIME coefficient and occurrence count for Anchors, both out of $100$ runs of the default algorithms. 

\subsection{Qualitative evaluation}
\label{sec:qualitative}
\subsubsection{Simple decision trees.}
\label{sec:dtree}
We first focus on simple decision trees relying on the presence or absence of given words. 
Such rules can be written in terms of indicator functions. 
We present four cases of increasing complexity. 

\begin{figure}[b]
    \centering
    \includegraphics[scale=0.35]{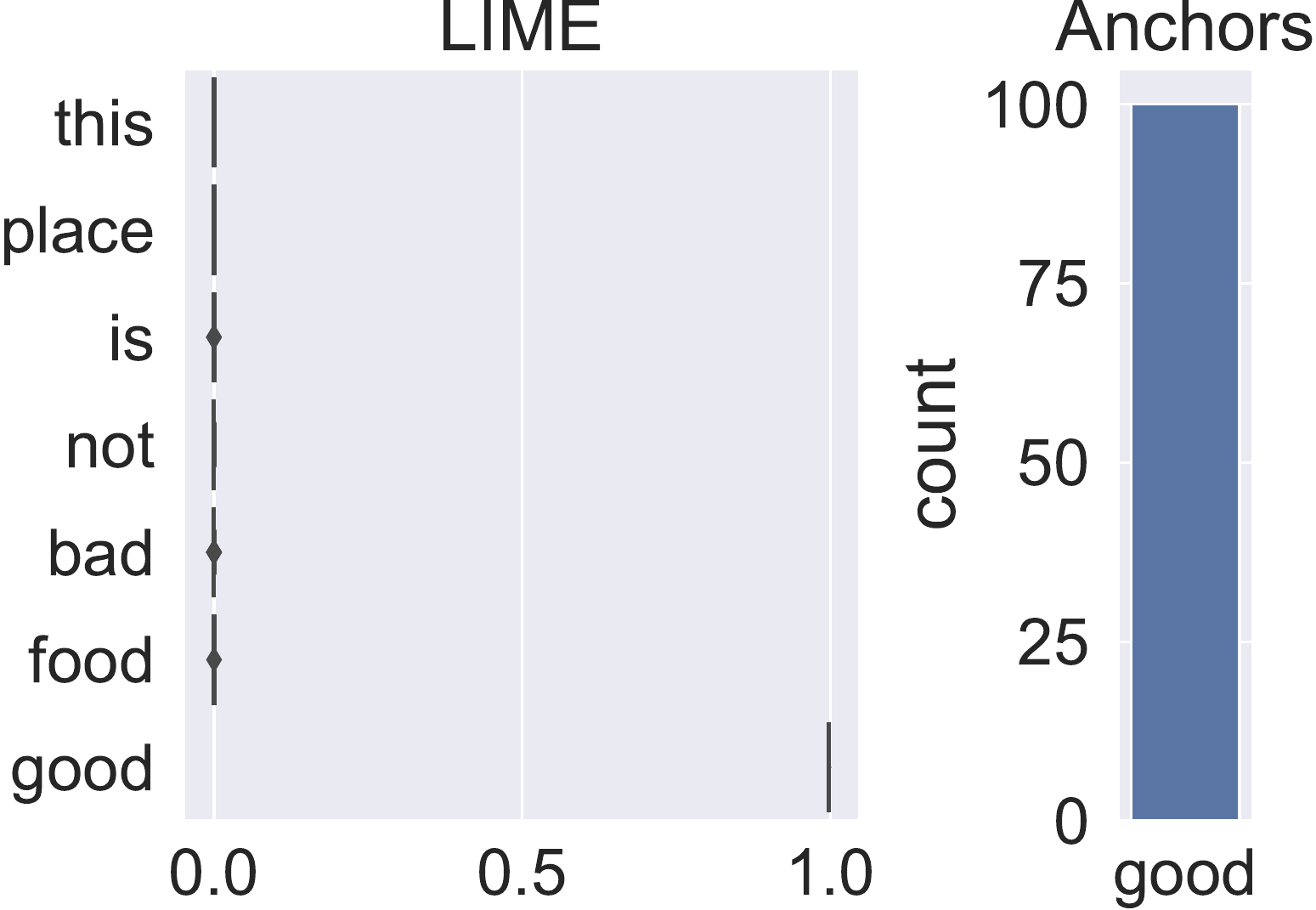}
    \hfill
    \includegraphics[scale=0.35]{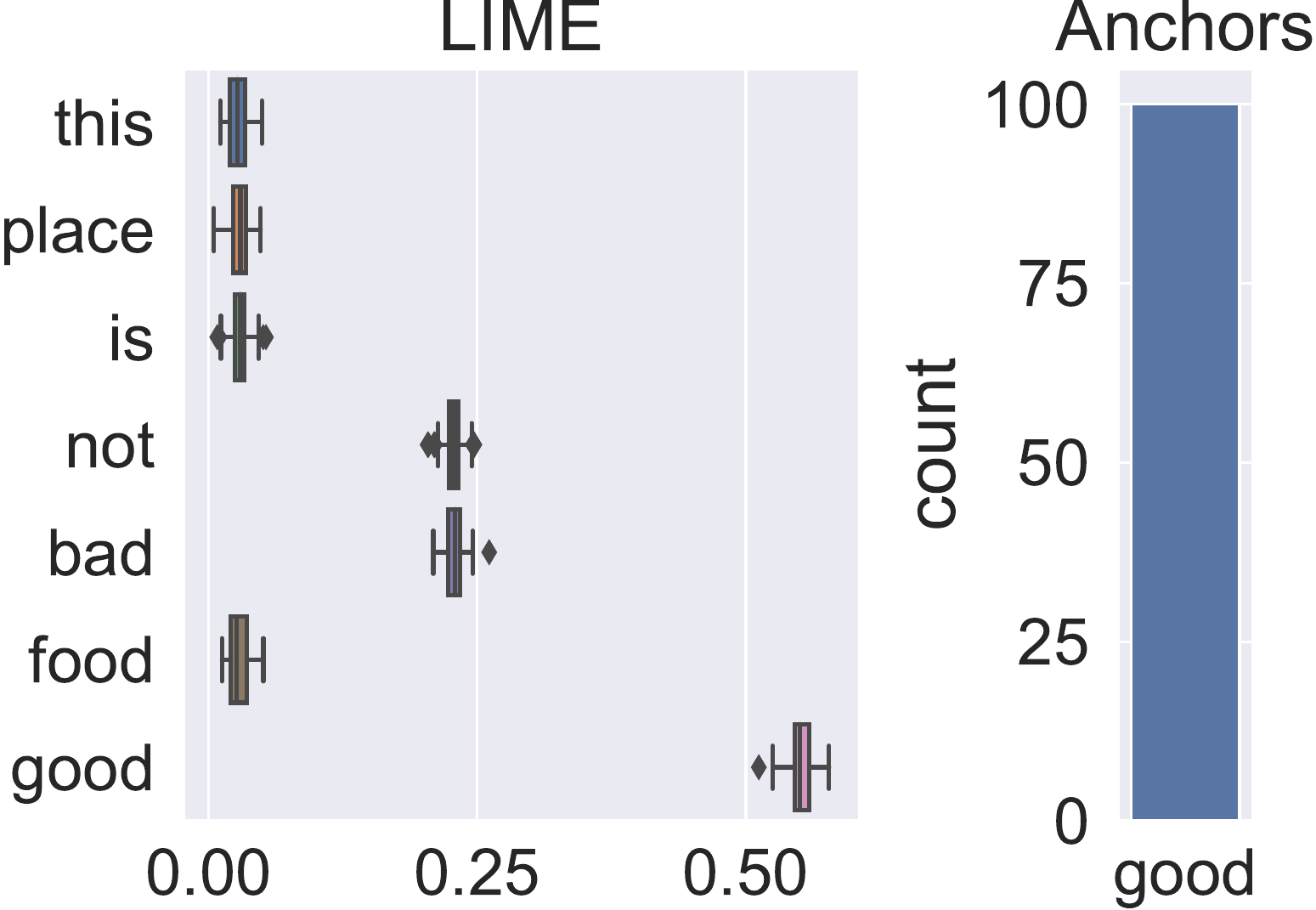}
\vspace{-0.2cm}
    \caption{\label{fig:dtree} Comparison on the classifiers $\indicator_{{\text{good}}\in \doc}$ (left panel) and $\indicator_{({\text{not}}\in \doc \;\text{and}\; {\text{bad}}\in \doc) \;\text{or}\; {\text{good}}\in \doc}$ (right panel) applied to the same review. Anchors makes no difference between the two.}
\end{figure}

\paragraph{Presence of a given word.}
Let us first look into the case of a simple decision tree returning $1$ or $0$ according to the presence or the absence of an individual word $\word_j\in\dictionary$, \emph{i.e.}, $f(\doc) = \indicator_{\word_j\in \doc} = \indicator_{\phi(\doc)_j>0}$. 
Let us consider an example $\xi$ such that $\word_j\in\xi$, meaning $f(\xi)=1$. 
In this case, \textbf{both methods behave as expected}: LIME attributes high weight to $\word_j$ and negligible weight to the others words, while Anchors extracts the anchor $A=\{\word_j\}$, as showcased in Figure~\ref{fig:dtree}. 

\paragraph{Small decision tree.}
Let us consider now a small decision tree looking for the presence of the words $\word_1$ and $\word_2$ or word $\word_3$, \emph{i.e.}, 
\[
    f(\doc) = \indicator_{(\word_1\in \doc \;\text{and}\; \word_2\in \doc) \;\text{or}\; \word_3\in x} \,. 
\]
We consider an example $\xi$ such that $\word_1,\word_2,\word_3\in\xi$. 
LIME assigns the same positive weight to $\word_1$ and $\word_2$, a higher weight to $\word_3$ and negligible weight to all other words, as shown in \citet{mardaoui2021analysis}. 
Anchors only extracts the word $\word_3$. 
In principle, we would expect the two methods to highlight the same words: they all seem important for the decision. 
Nevertheless, 
\textbf{Anchors is not considering $\word_1$ and $\word_2$ in its explanation, since the presence of word $\word_3$ is \textit{sufficient} to have a positive classification and $\{\word_3\}$ is a shorter anchor than $\{\word_1,\word_2\}$.}

\paragraph{Presence of several words.}
\begin{figure}[t]
    \centering
    \includegraphics[scale=0.35]{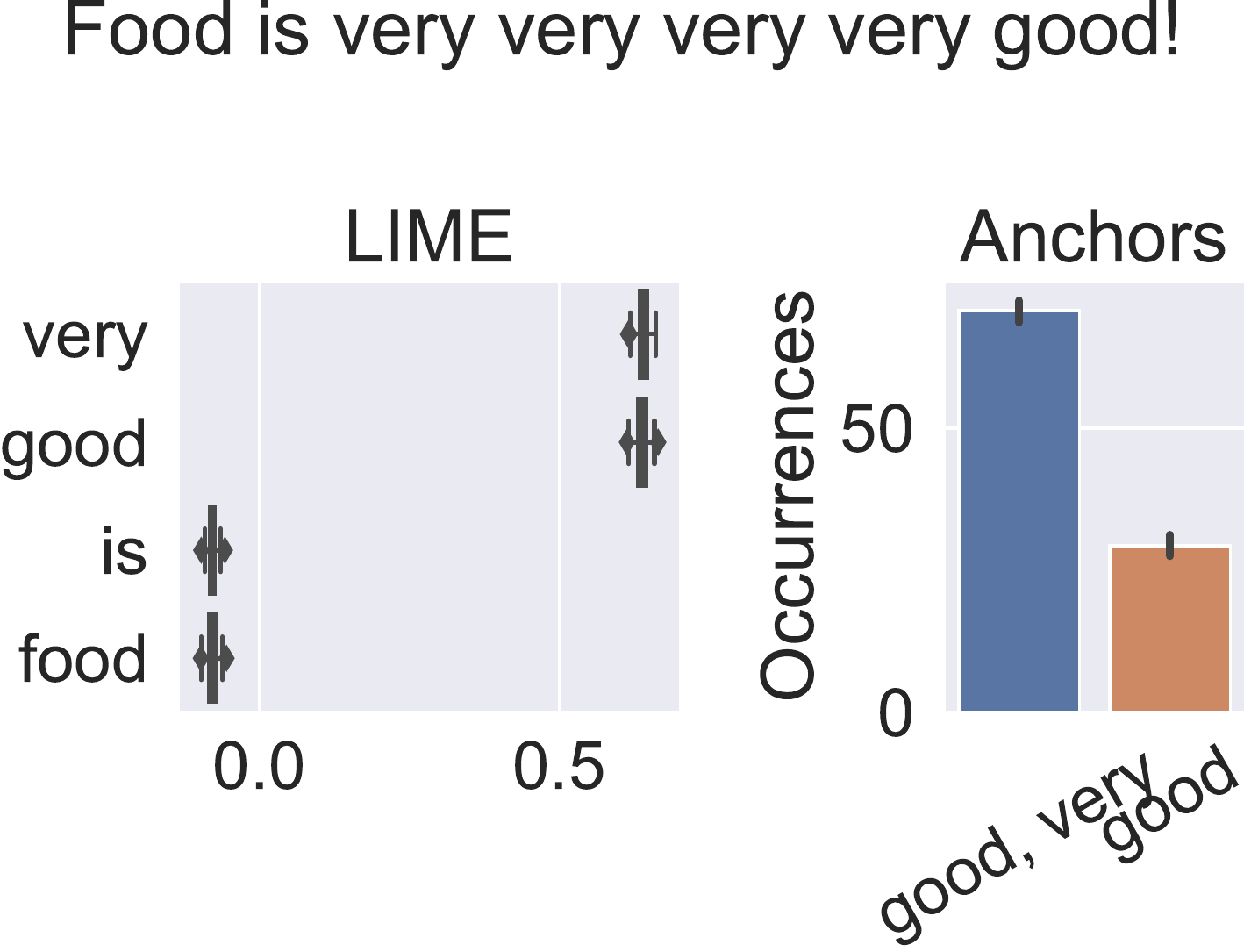}
    \hfill
    \includegraphics[scale=0.35]{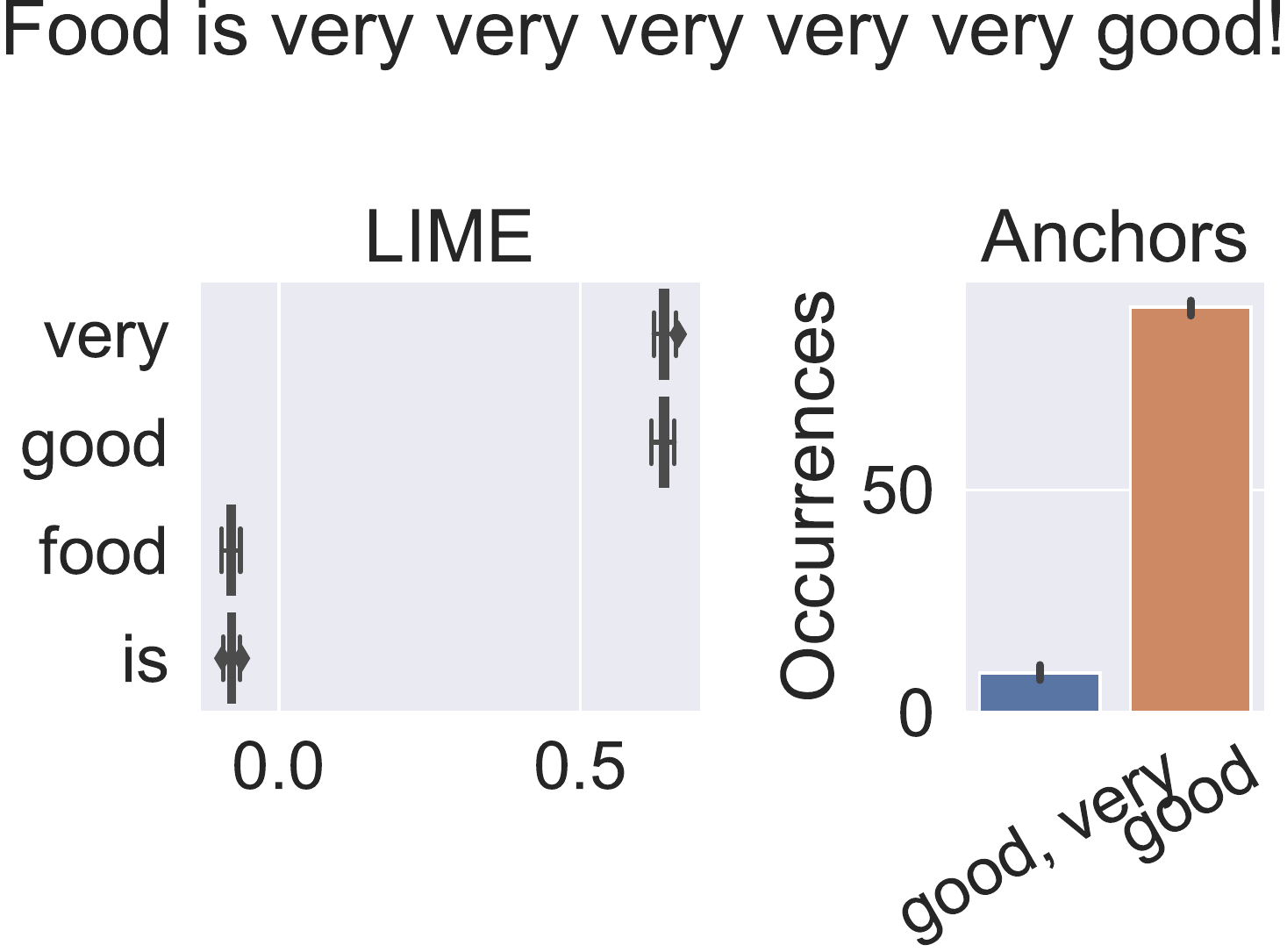}
\vspace{-0.3cm}
    \caption{\label{fig:product}Making a word disappear from the explanation by adding one occurrence. The classifier $\indicator_{({\text{very}}\in \doc \;\text{and}\; {\text{good}}\in \doc)}$ is applied when $\mult_{\text{very}}=4$ (left) and $\mult_{\text{very}}=5$ (right).}
\end{figure}
Let us generalize the previous example by considering 
a model classifying documents according to the presence or absence of a set of words.
Let $J=[k]\subseteq [d]$ be a set of distinct indices. 
We consider the model
\[
    f(\doc) = \prod_{j\in J} \indicator_{\word_j\in \doc} = \prod_{j\in J} \indicator_{\phi(\doc)_j>0} \,.
\]
Then LIME will assign the same importance to any word in $J$, independently from their multiplicities (Proposition~3 in \citet{mardaoui2021analysis}). 
On the contrary, Anchors explanations are impacted by the multiplicities of words (Proposition 6 in \citet{lopardo2022sea}). 
In particular, \textbf{if the multiplicity of a word in $J$ crosses a certain threshold, it disappears from the anchors} (see Figure~\ref{fig:product}). 
This is quite surprising, and not a desired behavior (especially since we do not control this threshold).

\paragraph{Presence of disjoint subsets of words.}
\begin{figure}[t]
    \centering
    \includegraphics[scale=0.35]{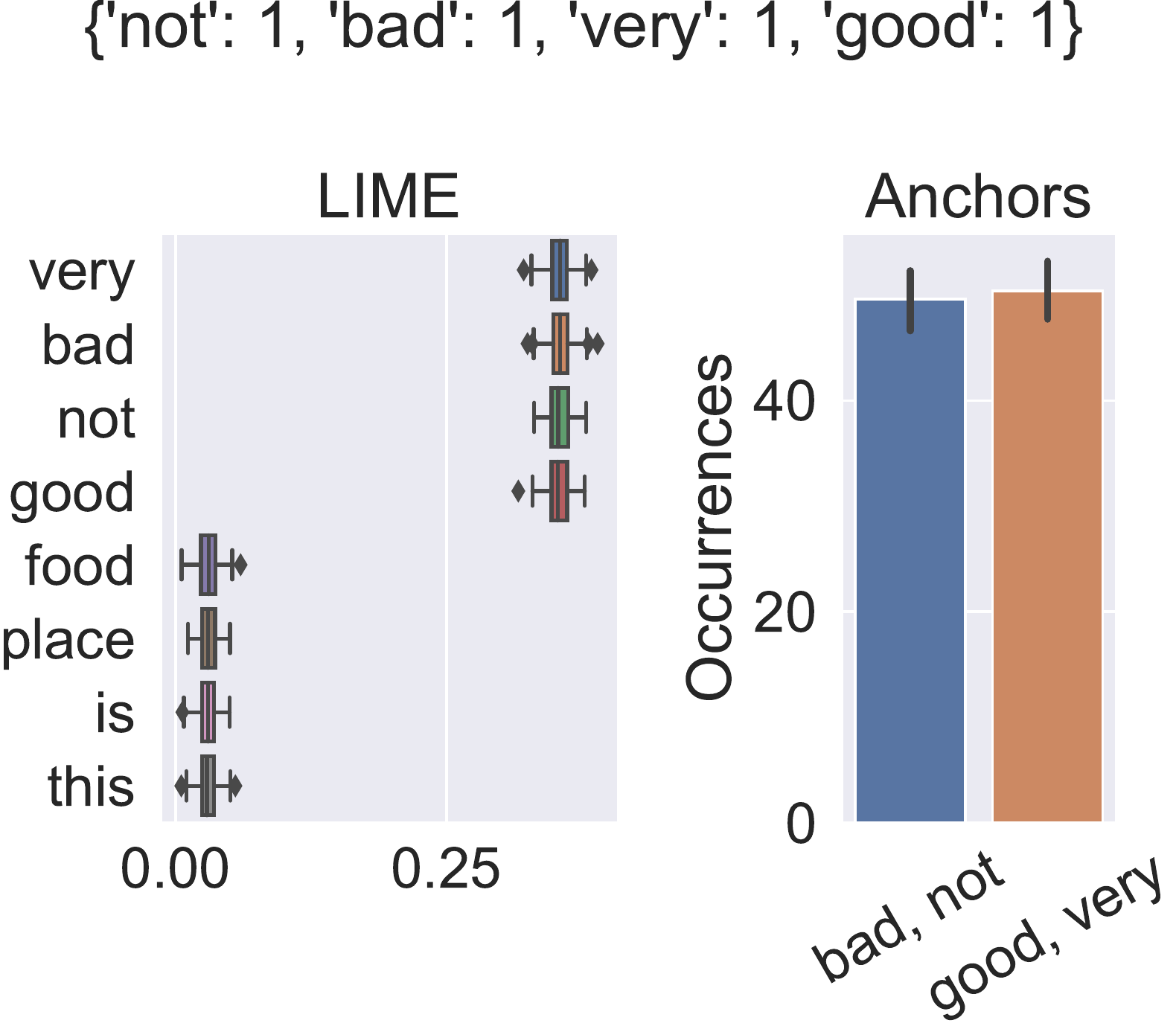}
    \hfill
    \includegraphics[scale=0.35]{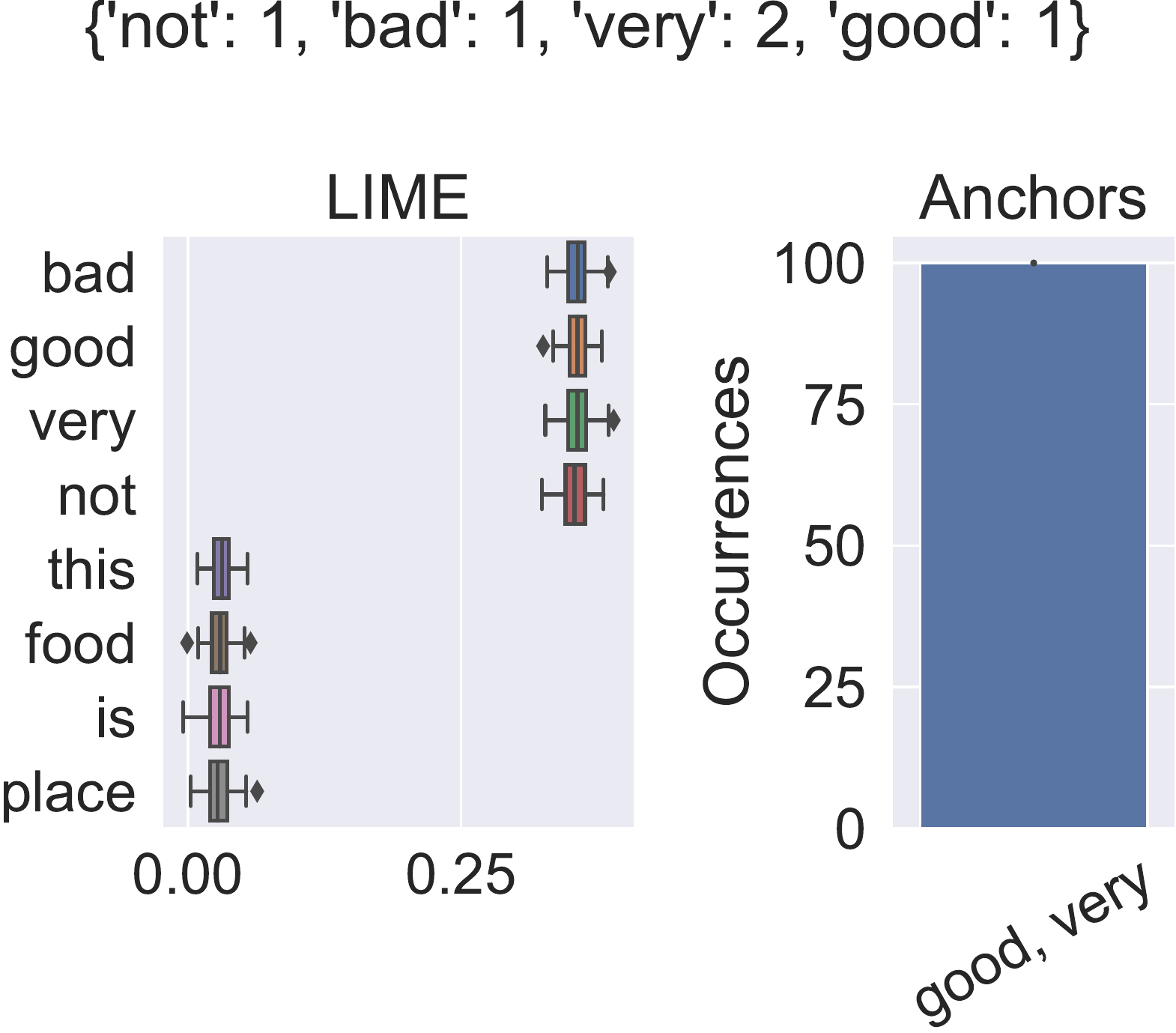}
    \medskip \\
    \includegraphics[scale=0.35]{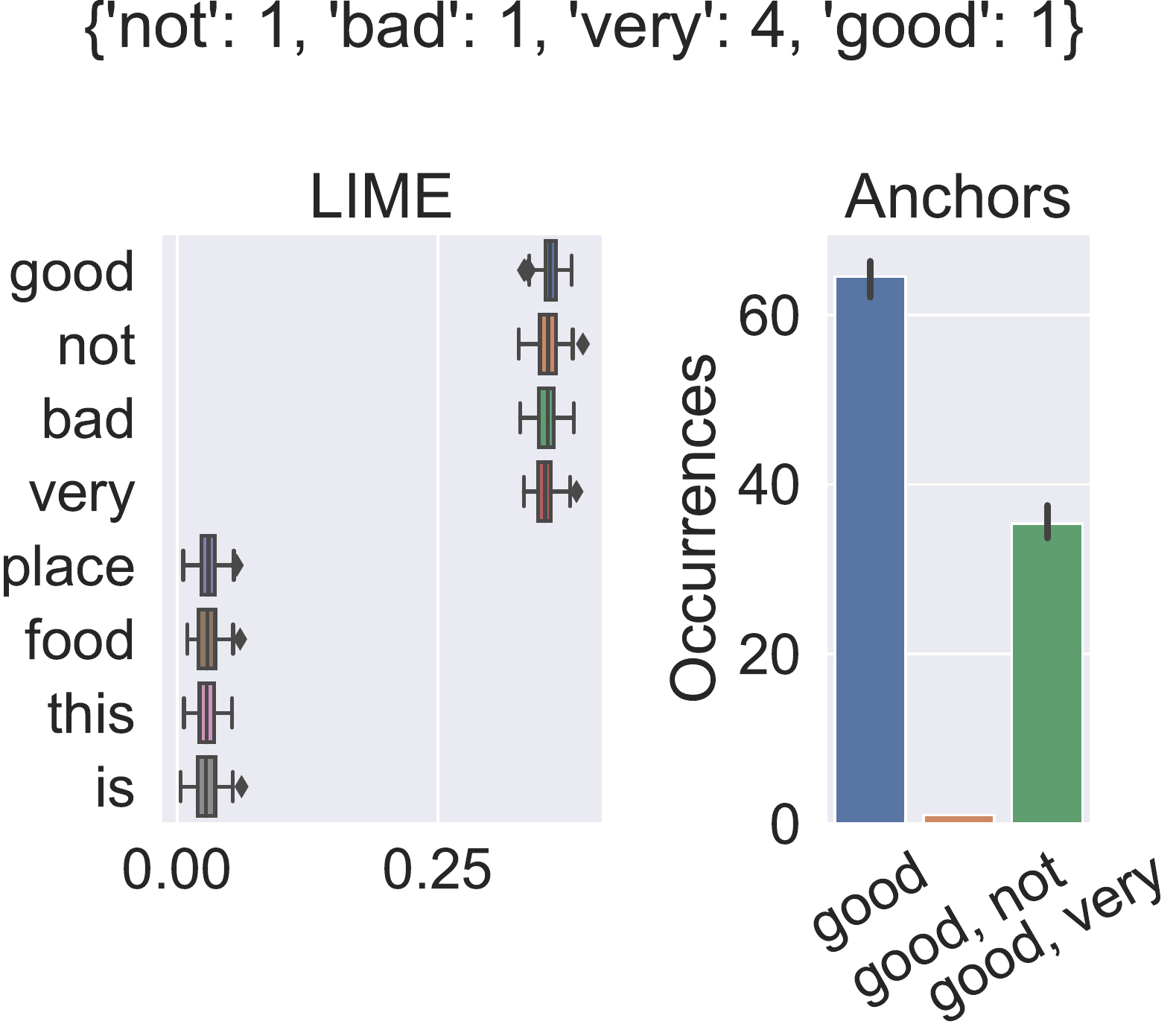}
    \hfill
    \includegraphics[scale=0.35]{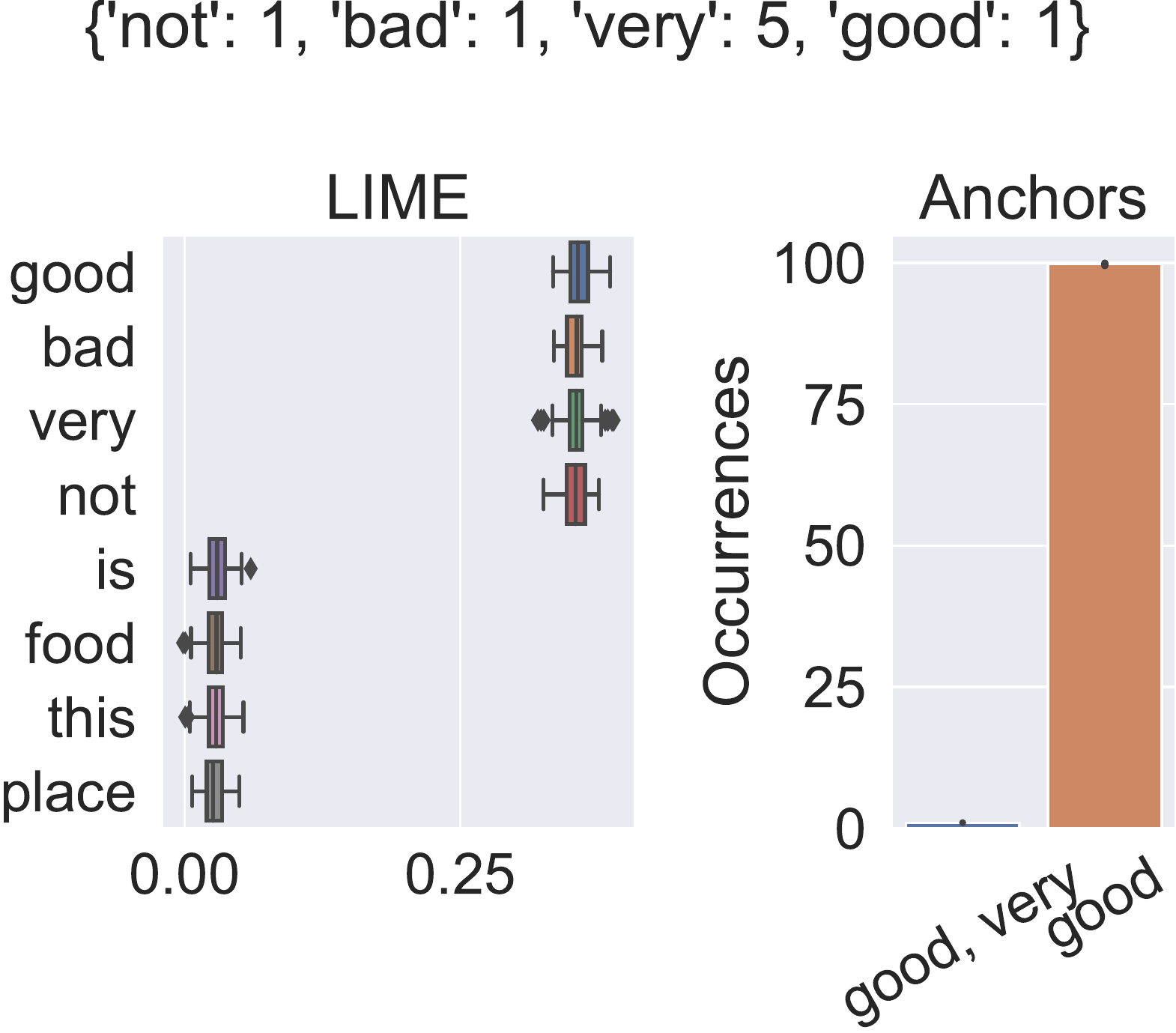}
\vspace{-0.3cm}
    \caption{\label{fig:subsets}Comparison on the classifier $\indicator_{({\text{not}}\in \doc \;\text{and}\; {\text{bad}}\in \doc) \;\text{or}\; ({\text{very}}\in \doc \;\text{and}\; {\text{good}}\in \doc)}$ when only $\mult_{\text{very}}$ is changing. Anchors' explanations depend on multiplicities.}
\end{figure}
Let us consider now two disjoint sets of indices $J_1=[k_1]\subseteq [d]$ and $J_2=\{k_1+1,\ldots,k_2\}\subseteq [d]$ with the same cardinality $\card{J_1}=\card{J_2}$. 
We consider the model
\begin{align*}
    f(\doc) 
         & = \prod_{j\in J_1} \indicator_{\word_j\in \doc} \cdot \prod_{j\in J_2} \indicator_{\word_j\in \doc}
         = \prod_{j\in J_1} \indicator_{\phi(\doc)_j>0} \cdot \prod_{j\in J_2} \indicator_{\phi(\doc)_j>0} \,,
\end{align*}
and an example $\xi$ such that $\word_j\in\xi$ for all $j\in J_1$ and for all $j\in J_2$.
LIME gives the same weight to words in $J_1$ and words in $J_2$. 
Anchors' explanations depend, again, on the multiplicities involved, (see Figure~\ref{fig:subsets}): as the occurrences of one word in $J_1$ (or in $J_2$) increase, the presence of other words in the same subset becomes \textit{sufficient} to get a positive prediction.


\subsubsection{Logistic models.}
\label{sec:logistic}
We now focus on logistic models.
Let $\sigma : \Reals \to [0,1]$ be the sigmoid function, that is, $t\mapsto 1/(1+e^{-t})$, $\lambda_0\in\Reals$ an intercept, and $\lambda\in\Reals^{d}$ fixed coefficients. 
Then, for any document $\doc$, we consider
    $f(\doc) = \indicator_{\sigma\left(\lambda_0+\lambda^\top\phi(\doc)\right)>\frac{1}{2}}.$

\begin{figure}[t]
    \centering
    \includegraphics[scale=0.35]{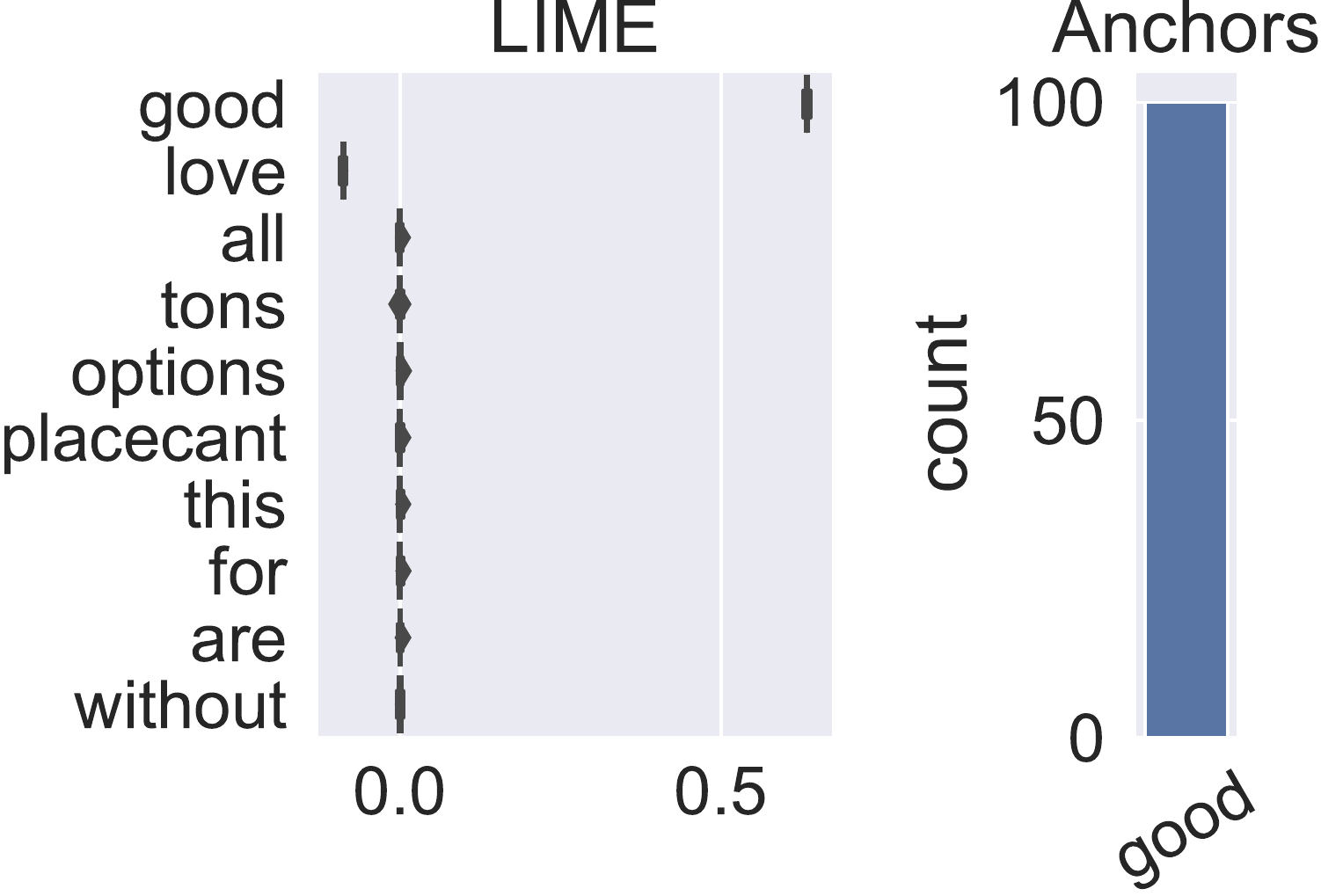}
    \hfill
    \includegraphics[scale=0.35]{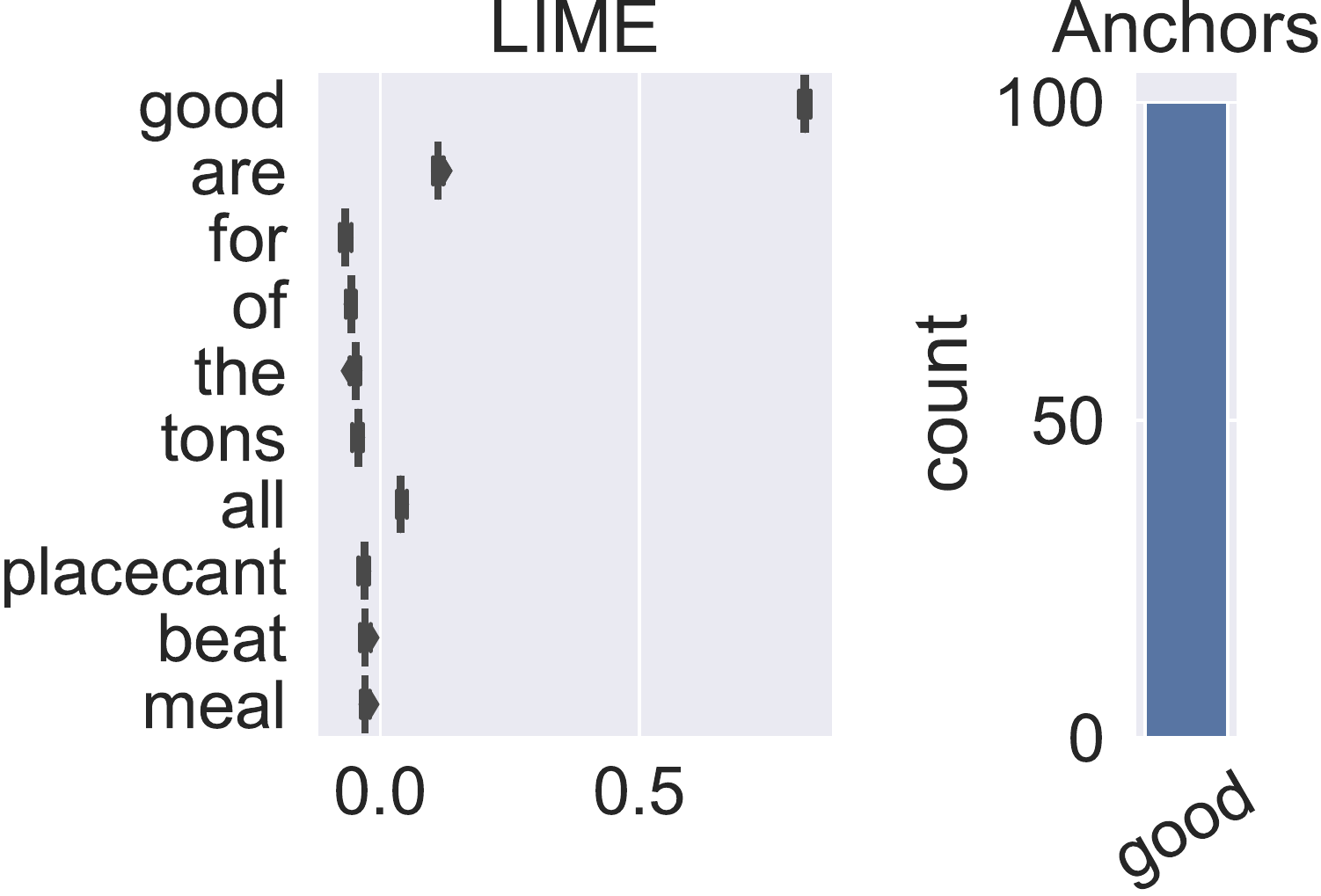}
\vspace{-0.3cm}
    \caption{\label{fig:logistic}Comparison on logistic model with $\lambda_{\text{love}}=-1$, $\lambda_{\text{good}}=+5$ and $\lambda_w=0$ for the others (left), \emph{vs} $\lambda_{\text{good}}=10$ and $\lambda_w\sim\mathcal{N}(0,1)$ for the others (right), applied to the same document. \emph{good} is the most important word for the classification in both cases.}
\end{figure}

\paragraph{Sparse case.}
We look at the case where only two coefficients $\lambda_1>0$ and $\lambda_2<0$ are nonzero, with $\abs{\lambda_1}>\abs{\lambda_2}$. 
LIME gives nonzero weights to $\word_1$ and $\word_2$, and null importance to the others, while Anchors only extract the word $\word_1$ (see Figure~\ref{fig:logistic}), as expected: $\word_1$ is the only word that really matters for positive prediction. 

\paragraph{Arbitrary coefficients.}
Let us take $\lambda_1\gg 0$, and $\lambda_j\sim\mathcal{N}(0,1)$, for $j \geq 2$.
Then LIME gives a weight $\lambda_1\gg 0$ and small weights to the others, while Anchors only extract $\{\word_1\}$: the only word actually influencing the decision (see Figure~\ref{fig:logistic}). 

When applied to simple if-then rules based on the presence of given words, we showed that Anchors has an unexpected behaviour with respect to the multiplicities of these words in a document, while LIME is perfectly capable of extracting the support of the classifier. 
The experiments on logistic models in Figure \ref{fig:logistic} show that, even when agreeing on the most important words, LIME is able to capture more information than Anchors.



\subsection{Quantitative evaluation}
\label{sec:quantitative}
When applying a logistic model $f$ on top of a vectorizer $\phi$, we know that the contribution of a word $\word_j$ is given by $\lambda_j\phi(\doc)_j$: we can unambiguously rank words in a document by importance. 
We propose to evaluate the ability of an explainer to detect the most important words for the classification of a document $\doc$ by measuring the similarity between the $N$ most important words for the interpretable classifier, namely $\logiwords_N(\doc)$, and the $N$ most important words according to the explainer, namely $E_N(\doc)$. 
We define the $\ell$-index for the explainer $E$ as
\[
   \ell_E \defeq \frac{1}{\abs{\corpus}} \sum_{\doc \in \corpus} \jaccard{E_N(\doc)}{\logiwords_N(\doc)}
    \,,
\]
where $\jaccard{\cdot}{\cdot}$ is Jaccard similarity and $\corpus$ is the test corpus.

Since we cannot fix $N$ a priori for Anchors, we run the experiments as follows. 
For any document $\doc$, we call $A(\doc)$ the obtained anchor and we use $N=\abs{A(\doc)}$, \emph{i.e.}, we compute $\jaccard{A(\doc)}{\logiwords_{\abs{A(\doc)}}(z)}$ for Anchors and $\jaccard{L_{\abs{A(\doc)}}(\doc)}{\logiwords_{\abs{A(\doc)}}(z)}$ for LIME.
Table \ref{tab:ell_results} shows the $\ell$-index and the computing time for LIME and Anchors on three different datasets. 
LIME has high performance in extracting the most important words, while requiring less computational time than Anchors. 
\textbf{An anchor is a minimal set of words that is sufficient (with high probability) to have a positive prediction, but it does not necessarily coincide with the $\abs{A}$ most important words for the prediction}.

\begin{table}
\caption{\label{tab:ell_results}Comparison between LIME and Anchors in terms of $\ell$-index and time.}
    \centering
    \begin{tabular}{c | c c c | c c c }
    & \multicolumn{3}{c}{$\ell$-index} & \multicolumn{3}{c}{time ($s$)} \\
     & Restaurants & Yelp & IMDB & Restaurants & Yelp & IMDB \\
    \hline
    LIME & $0.96 \pm 0.17$ & $0.95 \pm 0.22$ & $0.94 \pm 0.23$ & $0.21 \pm 0.05$ & $0.45 \pm 0.22$ & $0.73 \pm 0.44$ \\
    Anchors & $0.67 \pm 0.44$ & $0.29 \pm 0.43$ & $0.22 \pm 0.35$ & $0.19 \pm 0.27$ & $3.83 \pm 13.95$ & $33.87 \pm 165.08$ \\ 
    \end{tabular}
\end{table}



\section{Conclusion}
\label{sec:conclusion}
In this paper, we compared explanations on text data coming from two popular methods (LIME and Anchors), illustrating differences and unexpected behaviours when applied to simple models. 
We observe that the results can be quite different: the set of words $A$ extracted by Anchors does not coincide with the set of the $\abs{A}$ words with largest interpretable coefficients determined by LIME. 
We proposed the $\ell$-index to evaluate the ability of different explainers to identify the most important words. 
Our experiments show that LIME performs better than Anchors on this task, while requiring less computational resources.  

\paragraph{Acknowledgments.} Work supported by NIM-ML (ANR-21-CE23-0005-01).


\bibliography{biblio.bib}

\begin{thebibliography}{17}
\providecommand{\natexlab}[1]{#1}
\providecommand{\url}[1]{\texttt{#1}}
\expandafter\ifx\csname urlstyle\endcsname\relax
  \providecommand{\doi}[1]{doi: #1}\else
  \providecommand{\doi}{doi: \begingroup \urlstyle{rm}\Url}\fi

\bibitem[Ancona et~al.(2018)]{ancona_et_al_2018}
M.~Ancona et~al.
\newblock {Towards better understanding of gradient-based attribution methods
  for Deep Neural Networks}.
\newblock In \emph{ICLR}, 2018.

\bibitem[Bhatt et~al.(2021)Bhatt, Weller, and Moura]{bhatt2021evaluating}
U.~Bhatt, A.~Weller, and J.~Moura.
\newblock Evaluating and aggregating feature-based model explanations.
\newblock In \emph{IJCAI}, 2021.

\bibitem[Brown et~al.(2020)]{Brown_et_al_2020}
T.~Brown et~al.
\newblock Language models are few-shot learners.
\newblock \emph{{NeurIPS}}, 2020.

\bibitem[Devlin et~al.(2019)Devlin, Chang, Lee, and
  Toutanova]{devlin_et_al_2018}
J.~Devlin, M.~Chang, K.~Lee, and K.~Toutanova.
\newblock Bert: Pre-training of deep bidirectional transformers for language
  understanding.
\newblock \emph{NAACL}, 2019.

\bibitem[Guidotti et~al.(2018)Guidotti, Monreale, et~al.]{guidotti2018survey}
R.~Guidotti, A.~Monreale, et~al.
\newblock A survey of methods for explaining black box models.
\newblock \emph{ACM computing surveys (CSUR)}, 2018.

\bibitem[Lakkaraju et~al.(2016)Lakkaraju, Bach, and
  Leskovec]{lakkaraju2016interpretable}
H.~Lakkaraju, S.~Bach, and J.~Leskovec.
\newblock Interpretable decision sets: A joint framework for description and
  prediction.
\newblock In \emph{SIGKDD}, 2016.

\bibitem[Lim et~al.(2009)Lim, Dey, and Avrahami]{lim2009and}
B.~Lim, A.~Dey, and D.~Avrahami.
\newblock Why and why not explanations improve the intelligibility of
  context-aware intelligent systems.
\newblock In \emph{SIGCHI}, 2009.

\bibitem[Linardatos et~al.(2021)Linardatos, Papastefanopoulos, and
  Kotsiantis]{linardatos2021explainable}
P.~Linardatos, V.~Papastefanopoulos, and S.~Kotsiantis.
\newblock {Explainable AI: A Review of Machine Learning Interpretability
  Methods}.
\newblock \emph{Entropy}, 2021.

\bibitem[Lopardo et~al.(2022)Lopardo, Garreau, and Precioso]{lopardo2022sea}
G.~Lopardo, D.~Garreau, and F.~Precioso.
\newblock {A Sea of Words: An In-Depth Analysis of Anchors for Text Data}.
\newblock \emph{arXiv preprint arXiv:2205.13789}, 2022.

\bibitem[Luhn(1957)]{luhn1957statistical}
H.~P. Luhn.
\newblock A statistical approach to mechanized encoding and searching of
  literary information.
\newblock \emph{IBM Journal of research and development}, 1957.

\bibitem[Lundberg and Lee(2017)]{lundberg2017unified}
S.~M. Lundberg and S.~Lee.
\newblock {A Unified Approach to Interpreting Model Predictions}.
\newblock \emph{NeurIPS}, 2017.

\bibitem[Mardaoui and Garreau(2021)]{mardaoui2021analysis}
D.~Mardaoui and D.~Garreau.
\newblock An analysis of {LIME} for text data.
\newblock In \emph{AISTATS}. PMLR, 2021.

\bibitem[Margot and Luta(2021)]{margot2021new}
V.~Margot and G.~Luta.
\newblock {A New Method to Compare the Interpretability of Rule-based
  Algorithms}.
\newblock \emph{MDPI AI}, 2021.

\bibitem[Nguyen and Mart{\'\i}nez(2020)]{nguyen2020quantitative}
A.~Nguyen and M.~R. Mart{\'\i}nez.
\newblock On quantitative aspects of model interpretability.
\newblock \emph{arXiv preprint arXiv:2007.07584}, 2020.

\bibitem[Ribeiro et~al.(2016)Ribeiro, Singh, and Guestrin]{ribeiro2016should}
M.~T. Ribeiro, S.~Singh, and C.~Guestrin.
\newblock \say{{Why} should {I} trust you?}: {Explaining} the predictions of
  any classifier.
\newblock In \emph{ACM SIGKDD}, 2016.

\bibitem[Ribeiro et~al.(2018)Ribeiro, Singh, and Guestrin]{ribeiro2018anchors}
M.~T. Ribeiro, S.~Singh, and C.~Guestrin.
\newblock {Anchors: High-precision model-agnostic explanations}.
\newblock In \emph{{AAAI}}, 2018.

\bibitem[Vaswani et~al.(2017)]{vaswani_et_al_2017}
A.~Vaswani et~al.
\newblock Attention is all you need.
\newblock In \emph{NeurIPS}, 2017.

\end{thebibliography}

\end{document}